\documentclass[letterpaper, 10 pt, conference]{ieeeconf}  

\IEEEoverridecommandlockouts                              

\usepackage{graphics} 
\usepackage{epsfig} 
\usepackage{mathptmx} 
\usepackage{times} 
\usepackage{amsmath} 
\usepackage{amssymb}
\usepackage{bm}
\usepackage{algorithm}
\usepackage{algpseudocode}
\usepackage{hyperref}
\hypersetup{
  colorlinks = true, 
  urlcolor = black, 
  linkcolor = black, 
  citecolor = black 
}
\usepackage{multirow}
\usepackage{graphicx}

\title{\LARGE \bf
Real-Time Non-Contact Force Compensation for Wrist-Mounted Force/Torque Sensors in Haptic-Enabled Robotic Surgery Training*
}

\author{Walid Shaker and Mustafa Suphi Erden
\thanks{*This study was partially funded by the Engineering and Physical Sciences Research Council of the UK, under the EPSRC Grant EP/Y017307/1.}
\thanks{Walid Shaker and Mustafa Suphi Erden are with the School of Engineering and Physical Sciences, Heriot-Watt University, Edinburgh EH14 4AS, UK
{\tt\small (wkhs2000@hw.ac.uk; m.s.erden@hw.ac.uk)}}
}

\begin{document}
\maketitle
\thispagestyle{empty}
\pagestyle{empty}

\begin{abstract}

Haptic feedback has been a long-missed feature in robotic-assisted surgery, one that would allow surgeons to perceive tissue properties and apply controlled forces during delicate procedures. Although commercial robotic systems have begun to integrate haptic technologies, their high costs limit accessibility for training and research purposes. To address this gap, we extend our previously developed low-cost robotic surgery training setup, RoboScope, by incorporating a wrist-mounted force/torque (F/T) sensor for haptic feedback training. Wrist-mounted sensing avoids many challenges associated with tip-mounted sensors but introduces additional non-contact forces, such as gravity, sensor bias, installation offsets, and associated torques, which compromise measurement accuracy. In this paper, we propose a robust real-time compensation method based on recursive least squares (RLS). This method eliminates the need for dataset collection and frequent recalibration while adapting to changing operating conditions. Experimental validation demonstrates that the proposed approach achieves over 95\% error reduction in non-contact force compensation and more than 91\% in non-contact torque compensation, significantly outperforming existing methods. These results highlight the potential of our approach for providing reliable haptic feedback in robotic surgery training and research. 

\end{abstract}

\section{INTRODUCTION}

Haptic feedback is a critical feature in robotic-assisted surgery, as it enables surgeons to more effectively perceive tissue properties and regulate applied forces during delicate procedures \cite{gultom2025implementation}. Recent advances in surgical robotics, such as the da Vinci 5 system (Intuitive Surgical) and the Senhance® Surgical System (Asensus Surgical), have introduced haptic feedback into clinical practice, though only in a limited capacity. In addition, the high costs of these platforms constrain their accessibility, particularly for training and research contexts. To address this gap, we developed RoboScope \cite{shaker2025developing}-\cite{trute2022development}, a low-cost robotic surgery training setup designed to provide accessible, realistic, and remote training opportunities.

In this work, we extend RoboScope by integrating force-sensing capabilities to provide haptic feedback, helping trainees develop skills aligned with the latest advancements in robot-assisted surgery. Specifically, we focus on a wrist-mounted F/T sensor, which avoids many of the integration and durability challenges of tip-mounted sensors, simplifies maintenance, and eliminates the need to embed expensive components in potentially disposable or single-use surgical instruments. However, wrist-mounted sensors do not directly capture tip–tissue interaction forces, and their measurements are influenced by additional effects that must be carefully compensated for.

\section{BACKGROUND AND LITERATURE REVIEW}

Installing a six-axis F/T sensor on the robot's wrist enables force control, where accurately measuring interaction forces between the robot and its environment is critical, such as in assembly, polishing, milling, and physical human–robot interaction tasks. Nevertheless, the F/T sensor readings capture not only the actual external contact forces, but also internal non-contact forces generated by the robot motion, e.g., gravity, inertial, and Coriolis/centrifugal forces, as well as their associated torques. In addition, the sensor mechanical error, particularly its zero offset, can introduce further inaccuracies that must not be overlooked \cite{yu2021bias}-\cite{huang2024offsets}.

In low-speed applications, such as high-precision surface finishing, human–robot interaction, and haptic systems, inertial and Coriolis/centrifugal effects are typically negligible. However, real-time compensation for gravitational forces and torques remains essential, as relying on a static bias calibration before each execution is inadequate. While the gravity compensation is possible when the end-effector mass and centroid are known, in most practical scenarios, these parameters are unknown and must be identified. To perform gravity compensation effectively, it is also necessary to determine the transformation between the sensor frame and the robot end frame (sensor installation offset), as well as the orientation between the robot base frame and the gravity frame (robot installation offset) \cite{yu2021bias}-\cite{huang2024offsets}.

Several studies have been conducted to identify and compensate for these inaccuracies. Vougioukas \cite{vougioukas2001bias} employed least-squares estimation to identify the sensor zero offset and the gravity parameters of the end-effector. However, it necessitates positioning the robot in multiple predefined orientations. Authors in \cite{wang2018research} improved on this by considering the robot installation offset, acknowledging that the Z-axis of the robot base frame may not be aligned with gravity acceleration direction. Yang et al. \cite{yang2019force} introduced a multiparameter model to estimate the end-effector gravity, the sensor bias, and the sensor installation offset, but only considered rotation about the Z-axis between the sensor frame and robot end, limiting its generality. 

Authors in \cite{yu2021bias} further advanced the field by estimating the full rotation offset between the sensor and robot end, using a Frobenius norm constraint for the rotation matrix. Yet, this mathematical constraint is not fully equivalent to the strict requirements for a valid rotation matrix, requiring prior knowledge of the robot installation offset. Moreover, their method relies on placing the robot in a large number of carefully selected orientations. More recently, Huang et al. \cite{huang2024offsets} introduced a more efficient technique based on Gröbner basis approach and quaternion parameterization, enabling simultaneous estimation of all parameters and offsets with only three measurements and without prior information of the robot installation angle. Yet, relying only on three measurements results in limited accuracy, which requires to further employ a nonlinear least-squares method to refine the estimation results through multiple sets of measurements. In addition to these model-based approaches, researchers in \cite{lin2018dynamic} have used machine learning techniques, such as back-propagation neural networks, for gravity compensation. While these methods can be effective, they typically require large datasets and do not provide explicit estimates of physical parameters.

However, our practical implementation of the aforementioned model-based methods revealed that the sensor offset can drift over time. This drift necessitates frequent recalibration, sometimes before each execution, to maintain measurement accuracy. Furthermore, any changes in the setup installation typically require full recalibration and new data collection, which is time-consuming, particularly for approaches that rely on datasets or specific orientation measurements. 

Therefore, in this article, we propose a robust real-time compensation method that compensates for the non-contact forces and torques associated with wrist-mounted sensors and eliminates the need for datasets and frequent recalibration, while remaining resilient to changing operating conditions. 

The remainder of this article is organized as follows: Section~\ref{sec:Problem} presents the problem formulation. Section~\ref{sec:Method} describes the proposed non-contact force compensation method. Section~\ref{sec:Results} details the experimental investigation used to evaluate the algorithm, including parameter identification, compensation under no contact, and validation under external loads. Finally, Section~\ref{sec:Conc} concludes the paper and discusses potential directions for future work.

\section{PROBLEM FORMULATION}
\label{sec:Problem}

\begin{figure}[t!]
\centering
\includegraphics[scale=0.44]{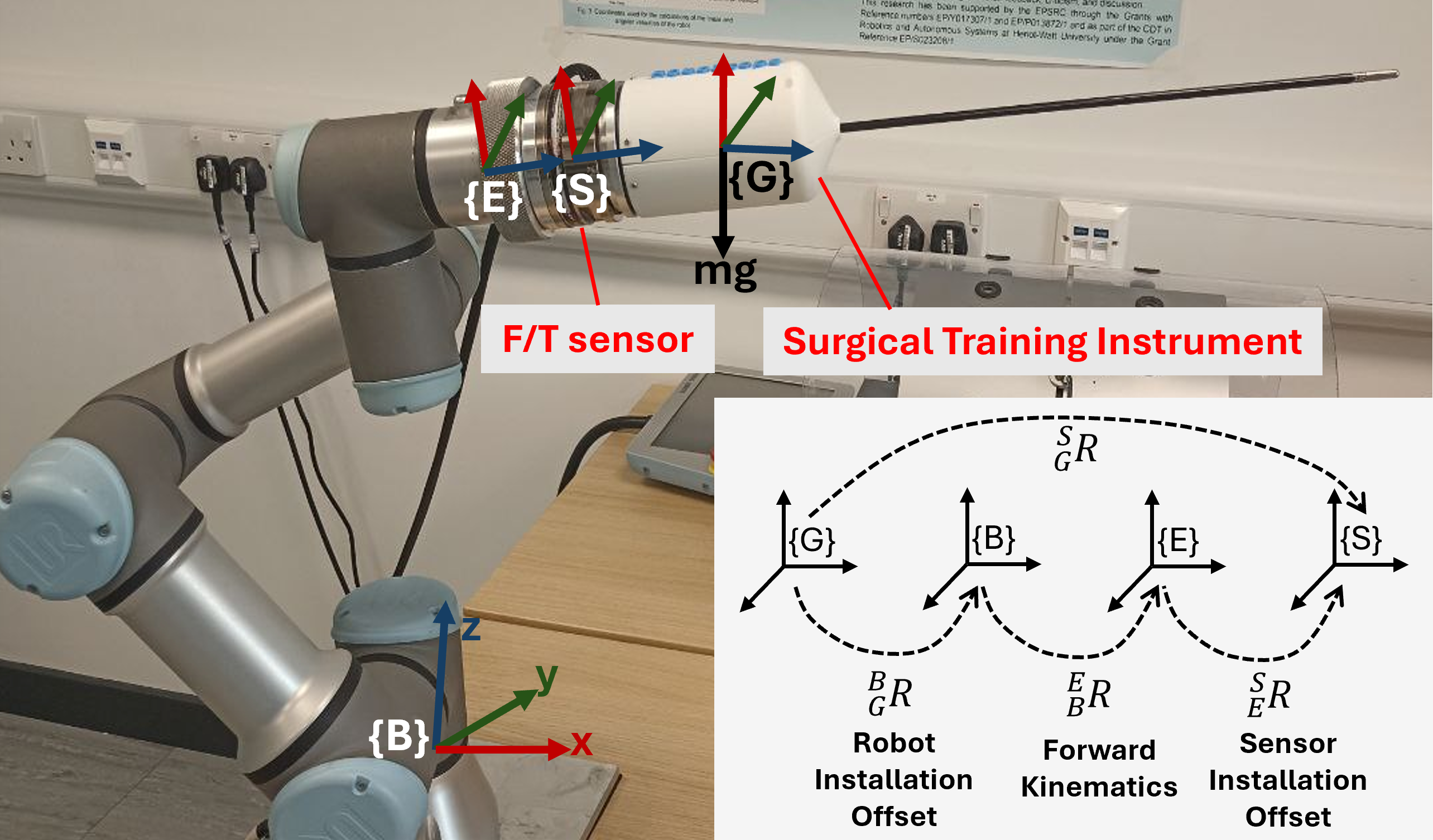}
\caption{The coordinate frames used for offsets estimation and gravity compensation in a wrist-mounted F/T sensor configuration. Commonly, the robot base frame is not aligned with the gravity reference frame.}
\label{fig:Frames}
\end{figure}

In this work, we consider a scenario where an ATI Gamma 6-axis F/T sensor is mounted at the robot's wrist and attached to a surgical training instrument, as illustrated in Fig.~\ref{fig:Frames}. To formulate the problem, the following coordinate frames are identified:
\begin{itemize}
    \item $\{B\}$: Base frame, attached to the robot base.
    \item $\{E\}$: Robot end frame, attached to the robot wrist.
    \item $\{S\}$: Sensor frame, attached to the F/T sensor.
    \item $\{G\}$: Gravity frame, arbitrarily attached to the world.
\end{itemize}

The raw force and torque readings of the sensor are represented by ${}^{S}\boldsymbol{f} \in \mathbb{R}^3$ and ${}^{S}\boldsymbol{\tau}  \in \mathbb{R}^3$, where

\begin{equation}
    {}^{S}\boldsymbol{f} = [{}^{S}{f}_{x}, {}^{S}{f}_{y}, {}^{S}{f}_{z}]^T 
    \label{eq:F}
\end{equation}
\begin{equation}
    {}^{S}\boldsymbol{\tau} = [{}^{S}{\tau}_{x}, {}^{S}{\tau}_{y}, {}^{S}{\tau}_{z}]^T
    \label{eq:T}
\end{equation}

\noindent where ${}^{S}{f}_{x}, {}^{S}{f}_{y}, {}^{S}{f}_{z}$ are the Cartesian force, ${}^{S}{\tau}_{x}, {}^{S}{\tau}_{y}, {}^{S}{\tau}_{z}$ are the torque components, and the superscript $T$ denotes the transpose of a matrix.

As explained earlier, the raw force and torque sensor measurements contain contact and non-contact components as

\begin{equation}
    {}^{S}\boldsymbol{f} = {}^{S}\boldsymbol{f}_{contact} + {}^{S}\boldsymbol{f}_{non\text{-}contact}
    \label{eq:F_raw}
\end{equation}
\begin{equation}
    {}^{S}\boldsymbol{\tau} = {}^{S}\boldsymbol{\tau}_{contact} + {}^{S}\boldsymbol{\tau}_{non\text{-}contact}
    \label{eq:T_raw}
\end{equation}

\noindent where ${}^{S}\boldsymbol{f}_{contact}$ and ${}^{S}\boldsymbol{\tau}_{contact}$ are the external actual forces and torques measured during contact. In contrast, the ${}^{S}\boldsymbol{f}_{non\text{-}contact}$ and ${}^{S}\boldsymbol{\tau}_{non\text{-}contact}$ denote the internal forces and torques that must be eliminated. These include the sensor zero offset, also known as sensor bias or systematic error (${}^{S}\boldsymbol{f}_{0}, {}^{S}\boldsymbol{\tau}_{0}$), as well as portions generated by robot motion such as gravitational forces (${}^{S}\boldsymbol{f}_{g}, {}^{S}\boldsymbol{\tau}_{g}$), inertial forces due to linear and angular acceleration (${}^{S}\boldsymbol{f}_{i}, {}^{S}\boldsymbol{\tau}_{i}$), and Coriolis/centrifugal forces (${}^{S}\boldsymbol{f}_{c}, {}^{S}\boldsymbol{\tau}_{c}$).

For low-speed scenarios, inertial forces and Coriolis/centrifugal forces can be neglected, i.e.,

\begin{equation}
    {}^{S}\boldsymbol{f}_{i} \approx 0, \quad {}^{S}\boldsymbol{\tau}_{i} \approx 0, \quad {}^{S}\boldsymbol{f}_{c} \approx 0, \quad {}^{S}\boldsymbol{\tau}_{c} \approx 0
    \label{eq:Assumption}
\end{equation}

\noindent Then (\ref{eq:F_raw}) and (\ref{eq:T_raw}) can be reformulated as

\begin{equation}
    {}^{S}\boldsymbol{f} = {}^{S}\boldsymbol{f}_{contact} + {}^{S}\boldsymbol{f}_{g} + {}^{S}\boldsymbol{f}_{0}
    \label{eq:F_comp}
\end{equation}
\begin{equation}
    {}^{S}\boldsymbol{\tau} = {}^{S}\boldsymbol{\tau}_{contact} + {}^{S}\boldsymbol{\tau}_{g} + {}^{S}\boldsymbol{\tau}_{0}
    \label{eq:T_comp}
\end{equation}

According to Fig.~\ref{fig:Frames}, the ${}^{S}\boldsymbol{f}_{g}, {}^{S}\boldsymbol{\tau}_{g}$ represents the force and torque in the sensor frame caused by the gravity of the end-effector, including the instrument and any attached load. They depend on the following

\begin{itemize}
    \item The instrument mass $m$
    \item The local gravity acceleration $g$
    \item The instrument center of gravity in sensor frame ${}^{S}_{G}\boldsymbol{p}$
    \item The orientation of gravity frame with respect to sensor frame ${}^{S}_{G}\boldsymbol{R} \in \mathbb{SO}(3)$, where $\mathbb{SO}(3)$ is the special orthogonal group, $\mathbb{SO}(3) = \{\mathbf{R} \in \mathbb{R}^{3 \times 3} \;|\; \mathbf{R} \cdot \mathbf{R}^{\mathrm{T}} = \mathbf{R}^{\mathrm{T}} \cdot \mathbf{R} = \mathbf{I}, \; det \;\mathbf{R} = 1\}$
\end{itemize}

The gravity frame $\{$G$\}$ can be defined arbitrarily; thus, its origin can be set at the centroid of the instrument. In this configuration, the force and torque generated by the gravity of the instrument, expressed in the gravity frame, are denoted as $\boldsymbol{f}_{mg}$ and $\boldsymbol{\tau}_{mg}$, respectively, where $\boldsymbol{f}_{mg} = [0, 0, -mg]^T$ and $\boldsymbol{\tau}_{mg} = [0, 0, 0]^T$. These can be transformed into the sensor frame as ${}^{S}\boldsymbol{f_}{g}$ and ${}^{S}\boldsymbol{\tau}_{g}$, and combined into a single gravitational wrench denoted by

\begin{equation}
\begin{bmatrix}
{}^{S}\boldsymbol{f}_{g} \\
{}^{S}\boldsymbol{\tau}_{g} 
\end{bmatrix}
=
\begin{bmatrix}
{}^{S}_{G}\boldsymbol{R} & \boldsymbol{0}_{3 \times 3} \\
{}^{S}_{G}\hat{\boldsymbol{p}} \cdot {}^{S}_{G}\boldsymbol{R} & {}^{S}_{G}\boldsymbol{R}
\end{bmatrix}
\begin{bmatrix}
\boldsymbol{f}_{mg} \\
\boldsymbol{0}_{3 \times 1}
\end{bmatrix}
\label{eq:gravity_wrench}
\end{equation}

\noindent where ${}^{S}_{G}\hat{\boldsymbol{p}} \in \mathbb{SO}(3)$ denotes the skew-symmetric matrix corresponding to the position vector ${}^{S}_{G}\boldsymbol{p} = [p_{x},\, p_{y},\, p_{z}]^T \in \mathbb{R}^3$. Accordingly, ${}^{S}_{G}\hat{\boldsymbol{p}}$ is given by

\begin{equation}
{}^{S}_{G}\hat{\boldsymbol{p}}
=
\begin{bmatrix}
0 & -p_{z} & p_{y} \\
p_{z} & 0 & -p_{x} \\
-p_{y} & p_{x} & 0 
\end{bmatrix}
\label{eq:P_sg}
\end{equation}

Since ${}^{S}_{G}\boldsymbol{R}$ is unknown, it can be obtained indirectly by the following transformation chain, as illustrated in Fig.~\ref{fig:Frames}.

\begin{equation}
{}^{S}_{G}\boldsymbol{R}
=
{}^{S}_{E}\boldsymbol{R} \cdot {}^{E}_{B}\boldsymbol{R} \cdot {}^{B}_{G}\boldsymbol{R} 
\label{eq:R_sg}
\end{equation}

\noindent where 
\begin{itemize}
    \item ${}^{S}_{E}\boldsymbol{R} \in \mathbb{SO}(3)$ is the orientation of $\{$E$\}$ with respect to $\{$S$\}$, known for the sensor installation offset.
    \item ${}^{E}_{B}\boldsymbol{R} \in \mathbb{SO}(3)$ is the orientation of $\{$B$\}$ with respect to $\{$E$\}$, computed from the transpose of the rotational component of the robot forward kinematics.
    \item ${}^{B}_{G}\boldsymbol{R} \in \mathbb{SO}(3)$ is the orientation of $\{$G$\}$ with respect to $\{$B$\}$, representing the robot installation angle. This offset accounts for situations where the robot base is not aligned with the gravity direction due to the mechanical installation error. 
\end{itemize}

By substituting (\ref{eq:gravity_wrench}), (\ref{eq:R_sg}) into (\ref{eq:F_comp}), (\ref{eq:T_comp}), it leads to

\begin{equation}
    {}^{S}\boldsymbol{f} = {}^{S}\boldsymbol{f}_{contact} + {}^{S}_{E}\boldsymbol{R} \cdot {}^{E}_{B}\boldsymbol{R} \cdot {}^{B}_{G}\boldsymbol{R} \cdot \boldsymbol{f}_{mg} + {}^{S}\boldsymbol{f}_{0}
    \label{eq:F_expand}
\end{equation}
\begin{equation}
    {}^{S}\boldsymbol{\tau} = {}^{S}\boldsymbol{\tau}_{contact} + {}^{S}_{G}\hat{\boldsymbol{p}} \cdot {}^{S}_{E}\boldsymbol{R} \cdot {}^{E}_{B}\boldsymbol{R} \cdot {}^{B}_{G}\boldsymbol{R} \cdot \boldsymbol{f}_{mg} + {}^{S}\boldsymbol{\tau}_{0}
    \label{eq:T_expand}
\end{equation}

For simplicity, ${}^{B}_{G}\boldsymbol{R}$ and $\boldsymbol{f}_{mg}$ can be combined into a single term, introducing an unknown variable ${}^{B}\boldsymbol{f}$, as estimating them simultaneously is challenging \cite{yu2021bias}.

\begin{equation}
    {}^{B}\boldsymbol{f} = {}^{B}_{G}\boldsymbol{R} \cdot \boldsymbol{f}_{mg}
    \label{eq:Fb}
\end{equation}

\noindent where ${}^{B}\boldsymbol{f}$ denotes the gravitational force of the instrument expressed in the robot base frame. Using this substitution, (\ref{eq:F_expand}) and (\ref{eq:T_expand}) can be rewritten as 

\begin{equation}
    {}^{S}\boldsymbol{f} = {}^{S}\boldsymbol{f}_{contact} + {}^{S}_{E}\boldsymbol{R} \cdot {}^{E}_{B}\boldsymbol{R} \cdot {}^{B}\boldsymbol{f} + {}^{S}\boldsymbol{f}_{0}
    \label{eq:F_Fb}
\end{equation}
\begin{equation}
    {}^{S}\boldsymbol{\tau} = {}^{S}\boldsymbol{\tau}_{contact} + {}^{S}_{G}\hat{\boldsymbol{p}} \cdot {}^{S}_{E}\boldsymbol{R} \cdot {}^{E}_{B}\boldsymbol{R} \cdot {}^{B}\boldsymbol{f} + {}^{S}\boldsymbol{\tau}_{0}
    \label{eq:T_Fb}
\end{equation}

The unknown parameters that need to be identified are ${}^{S}_{E}\boldsymbol{R}$, ${}^{B}\boldsymbol{f}$, ${}^{S}\boldsymbol{f}_{0}$, ${}^{S}_{G}\hat{\boldsymbol{p}}$, ${}^{S}\boldsymbol{\tau}_{0}$. In the context of parameters estimation for wrist-mounted F/T sensors, there is no existing method capable of obtaining ${}^{S}_{E}\boldsymbol{R}$ and ${}^{B}\boldsymbol{f}$ simultaneously \cite{huang2024offsets}. Yu et al. \cite{yu2021bias} also indicated that due to the sensor installation offset ${}^{S}_{E}\boldsymbol{R}$, there is no direct closed-form solution using the least-squares method for the force compensation model (\ref{eq:F_Fb}). 

Since simultaneous identification of all model parameters is challenging, the sensor installation was designed to facilitate approximate alignment between the sensor frame $\{S\}$ and the robot end frame $\{E\}$. This design choice simplifies the estimation of the sensor installation rotation ${}^{S}_{E}\boldsymbol{R}$. Importantly, perfect alignment is not assumed. Instead, an independent validation procedure was performed to verify the residual rotation between the two frames. This procedure ensured that the remaining offset was negligible such that ${}^{S}_{E}\boldsymbol{R} \approx \boldsymbol{I}_{3 \times 3}$, and therefore did not affect the performance of the compensation method. 

If such an alignment or validation is not feasible, the sensor installation offset can be explicitly estimated. For example, Huang et al. \cite{huang2024offsets} demonstrated that ${}^{S}_{E}\boldsymbol{R}$ can be identified using a Gröbner basis approach. As a result, (\ref{eq:F_Fb}) and (\ref{eq:T_Fb}) simplify to

\begin{equation}
    {}^{S}\boldsymbol{f} = {}^{S}\boldsymbol{f}_{contact} + {}^{E}_{B}\boldsymbol{R} \cdot {}^{B}\boldsymbol{f} + {}^{S}\boldsymbol{f}_{0}
    \label{eq:F_no_Rse}
\end{equation}
\begin{equation}
    {}^{S}\boldsymbol{\tau} = {}^{S}\boldsymbol{\tau}_{contact} + {}^{S}_{G}\hat{\boldsymbol{p}} \cdot {}^{E}_{B}\boldsymbol{R} \cdot {}^{B}\boldsymbol{f} + {}^{S}\boldsymbol{\tau}_{0}
    \label{eq:T_no_Rse}
\end{equation}

To estimate the unknown parameters ${}^{B}\boldsymbol{f}$, ${}^{S}\boldsymbol{f}_{0}$, ${}^{S}_{G}\hat{\boldsymbol{p}}$, ${}^{S}\boldsymbol{\tau}_{0}$, and obtain the identification model, no external contact force or torque should be applied. Thus, ${}^{S}\boldsymbol{f}_{contact}$ = 0, ${}^{S}\boldsymbol{\tau}_{contact}$ = 0, resulting in 

\begin{equation}
    {}^{S}\boldsymbol{f} = {}^{E}_{B}\boldsymbol{R} \cdot {}^{B}\boldsymbol{f} + {}^{S}\boldsymbol{f}_{0}
    \label{eq:F_no_cont}
\end{equation}
\begin{equation}
    {}^{S}\boldsymbol{\tau} = {}^{S}_{G}\hat{\boldsymbol{p}} \cdot {}^{E}_{B}\boldsymbol{R} \cdot {}^{B}\boldsymbol{f} + {}^{S}\boldsymbol{\tau}_{0}
    \label{eq:T_no_cont}
\end{equation}

As (\ref{eq:F_no_cont}) and (\ref{eq:T_no_cont}) are valid when there is no contact, they can be used to eliminate the non-contact forces and estimate the actual contact force in case of a contact. In other words, once the unknown parameters are estimated, the contact force and torque can be derived as

\begin{equation}
    {}^{S}\boldsymbol{f}_{contact} = {}^{S}\boldsymbol{f} - {}^{E}_{B}\boldsymbol{R} \cdot {}^{B}\boldsymbol{f} - {}^{S}\boldsymbol{f}_{0}
    \label{eq:F_cont}
\end{equation}
\begin{equation}
    {}^{S}\boldsymbol{\tau}_{contact} = {}^{S}\boldsymbol{\tau} - {}^{S}_{G}\hat{\boldsymbol{p}} \cdot {}^{E}_{B}\boldsymbol{R} \cdot {}^{B}\boldsymbol{f} - {}^{S}\boldsymbol{\tau}_{0}
    \label{eq:T_cont}
\end{equation}

\section{METHOD DESCRIPTION}
\label{sec:Method}
 
The proposed method first estimates the actual contact force ${}^{S}\boldsymbol{f}_{contact}$ in Section~\ref{sec:force}, using the force compensation model (\ref{eq:F_no_cont}). In this step, the gravitational force in the robot base frame ${}^{B}\boldsymbol{f}$ and the sensor force bias ${}^{S}\boldsymbol{f}_{0}$ are identified from sensor measurements and robot kinematics. Subsequently, the actual contact torque ${}^{S}\boldsymbol{\tau}_{contact}$ is obtained using the torque compensation model (\ref{eq:T_no_cont}), in Section~\ref{sec:torque}, where the instrument center of gravity ${}^{S}_{G}\hat{\boldsymbol{p}}$ and the sensor torque bias ${}^{S}\boldsymbol{\tau}_{0}$ are estimated from sensor measurements and the parameters identified in the force compensation stage.

\subsection{Non-contact Force Compensation}
\label{sec:force} 

The force compensation model (\ref{eq:F_no_cont}) can be solved using least-squares estimation, where accurate parameter identification typically requires that the number of independent measurements ($n$) be significantly greater than the number of unknowns. With this in mind, (\ref{eq:F_no_cont}) can be reformulated as

\begin{equation}
    \begin{bmatrix}
        {}^{S}{f}_{1} \\
        {}^{S}{f}_{2} \\
        \vdots \\
        {}^{S}{f}_{n} 
    \end{bmatrix}
    =
    \begin{bmatrix}
        {}^{E}_{B}{R}_{1} & I_{3 \times 3} \\
        {}^{E}_{B}{R}_{2} & I_{3 \times 3} \\
        \vdots & \vdots \\
        {}^{E}_{B}{R}_{n}  & I_{3 \times 3} 
    \end{bmatrix}
    \cdot
    \begin{bmatrix}
        {}^{B}{f} \\
        {}^{S}{f}_{0} 
    \end{bmatrix}
    \label{eq:force_sys}
\end{equation}

\noindent The above equation can be rewritten as 

\begin{equation}
    {y} = {C}{x}
    \label{eq:y_Cx}
\end{equation}

\noindent where $y \in \mathbb{R}^{3n}$ is the sensor force measurement vector, $C \in \mathbb{R}^{3n \times 6}$ is the measurement matrix, and $x = \begin{bmatrix} {}^{B}{\boldsymbol{f}} & {}^{S}\boldsymbol{f}_{0} \end{bmatrix}^T \in \mathbb{R}^6$ is the vector of unknown parameters. To solve this system, the objective is to find an estimate $\hat{x}$ that minimizes the Euclidean norm of the residual $r = C\hat{x} - y$. This can be expressed as the following minimization problem.

\begin{equation}
    \min_x \left\lVert C\hat{x}-y \right\rVert^2_2
    \label{eq:min}
\end{equation}

As the minimum of $\left\lVert r \right\rVert^2_2$ coincides with the minimum of $(C\hat{x}-y)^T(C\hat{x}-y)$, the solution to this least-squares problem is given by 

\begin{align}
\begin{split}
    2C^{T}(C\hat{x}-y) = 0  \\
    C^{T}C\hat{x} = C^{T}y \\
    \hat{x} = (C^{T}C)^{-1}C^{T}y
    \label{eq:LSM}
\end{split}
\end{align}

The formulation and solution of the least-squares problem typically assume that all measurements are available at a certain time instant. In practice, this requires collecting all the data before solving the problem, which introduces latency. Furthermore, our practical implementation revealed that the sensor offset can drift over time, requiring repeated data collection and recalibration before each execution, an approach that is both time-consuming and inefficient. Therefore, it is essential to employ an adaptive approach that updates the estimation as new measurements arrive, enabling continuous calibration and reducing overall processing time.

These considerations motivate the use of recursive least squares (RLS) instead of the batch least-squares approach. RLS is an adaptive algorithm that efficiently updates the solution to a least-squares problem as new data becomes available, avoiding the need to recompute the solution from scratch. It is a widely used tool for online parameter estimation in systems that are linear in their parameters, as well as in adaptive filtering for signal processing and control. The RLS formulation adopted here follows the derivation in \cite{simon2006optimal}.

Mathematically, the RLS algorithm updates the estimate $\hat{x}_k \in \mathbb{R}^6$ at each discrete time instant $k$ by applying a correction to the previous estimate $\hat{x}_{k-1} \in \mathbb{R}^6$, leading to fast convergence to the optimal solution.

\begin{equation}
    \hat{x}_k = \hat{x}_{k-1} + K_k(y_k - C_k \hat{x}_{k-1})
    \label{eq:x_hat}
\end{equation}

\noindent Here, $y_k \in \mathbb{R}^3$ is the new sensor measurement, $K_k \in \mathbb{R}^{6 \times 3}$ is the gain matrix that determines how much the new measurement corrects the previous estimate, and $C_k \in \mathbb{R}^{3 \times 6}$ is the measurement matrix for the new sample. 

The gain matrix $K_k$ is obtained by minimizing the trace of the estimation error covariance matrix $P_k \in \mathbb{R}^{6 \times 6}$, yielding

\begin{equation}
    K_k = P_{k-1} C^T_k(R_k + C_k P_{k-1}C^T_k)^{-1}
    \label{eq:K_k}
\end{equation}

\noindent where $R_k$ denotes the covariance of the measurement noise and $P_{k-1}$ represents the prior estimation error covariance matrix. The propagation of the estimation error covariance matrix is given by

\begin{equation}
    P_k = (I - K_k C_k)P_{k-1} 
    \label{eq:P_k}
\end{equation}

If there is no prior knowledge of the unknown parameters, the algorithm is initialized with a random initial estimate $x_0$ and $P_0 = \infty I$, indicating a very large uncertainty. In contrast, if perfect knowledge of $x$ is available before any measurements are taken, then $P=0$ \cite{simon2006optimal}. Alternatively, $x_0$ and $P_0$ can be initialized with reasonable values from the previous sessions. The convergence to the optimal solution is defined by $\left\lVert \Delta{x} \right\rVert_2 < \epsilon$, where $\epsilon$ is the convergence threshold, and $\Delta{x}$ is given by 

\begin{equation}
    \Delta{x} = \hat{x}_{k} - \hat{x}_{k-1}
    \label{eq:dx}
\end{equation}

\noindent Once convergence is reached, ${}^{B}\boldsymbol{f}$ and ${}^{S}\boldsymbol{f}_{0}$ can be extracted from the vector $\hat{x}$, then using (\ref{eq:F_cont}), the actual external contact force ${}^{S}\boldsymbol{f}_{contact}$ can be computed. 

When contact occurs before convergence, it causes the RLS algorithm to interpret the contact forces as part of the bias and gravitational components, resulting in incorrect compensation. To mitigate this, a contact detection mechanism is employed to determine when to freeze or slow down the RLS updates. Contact is detected by monitoring the 2-norm of the residual $C\hat{x} - y$; if it exceeds a predefined threshold $f_{th}$, tuned experimentally, the measurement is interpreted as external contact and RLS updates are paused to prevent contamination of the estimates. When the residual falls below the threshold, updates resume automatically, enabling slow adaptation. The algorithm is initialized in a no-contact state to allow for rapid convergence of the model parameters, which is required before reliable contact forces can be extracted.


\subsection{Non-contact Torque Compensation}
\label{sec:torque} 

After compensating for the non-contact forces and obtaining ${}^{B}{\boldsymbol{f}}, {}^{S}{\boldsymbol{f}}_0$, the torque compensation model (\ref{eq:T_no_cont}) can be reformulated by substituting the term (${}^{E}_{B}\boldsymbol{R} \cdot {}^{B}\boldsymbol{f}$) from (\ref{eq:F_no_cont}). Then (\ref{eq:T_no_cont}) can be rewritten as 

\begin{equation}
    {}^{S}\boldsymbol{\tau} = {}^{S}_{G}\hat{\boldsymbol{p}} \cdot ({}^{S}{\boldsymbol{f}}-{}^{S}{\boldsymbol{f}}_0) + {}^{S}\boldsymbol{\tau}_{0}
    \label{eq:T_no_cont_edit}
\end{equation}

\noindent where the difference ${}^{S}{\boldsymbol{f}}-{}^{S}{\boldsymbol{f}}_0$ is denoted as ${\Delta \boldsymbol{f}}$. Thus, (\ref{eq:T_no_cont_edit}) can be expressed component-wise as

\begin{equation}
    \begin{bmatrix}
        {}^{S}\tau_x \\
        {}^{S}\tau_y \\
        {}^{S}\tau_z
    \end{bmatrix}
    =
    \begin{bmatrix}
        0 & -p_{z} & p_{y} \\
        p_{z} & 0 & -p_{x} \\
        -p_{y} & p_{x} & 0 
    \end{bmatrix}
    \cdot
    \begin{bmatrix}
        \Delta f_x \\
        \Delta f_y \\
        \Delta f_z
    \end{bmatrix}
    +
    \begin{bmatrix}
        {}^{S}\tau_{0,x} \\
        {}^{S}\tau_{0,y} \\
        {}^{S}\tau_{0,z} 
    \end{bmatrix}
    \label{eq:T_no_cont_sys}
\end{equation}

\noindent Then (\ref{eq:T_no_cont_sys}) can be written in the linear system form $y=Cx$ 

\begin{equation}
    \begin{bmatrix}
        {}^{S}\tau_x \\
        {}^{S}\tau_y \\
        {}^{S}\tau_z
    \end{bmatrix}
    =
    \begin{bmatrix}
        0 & \Delta f_z & -\Delta f_y  & 1 & 0 & 0\\
        -\Delta z & 0  & \Delta x & 0 & 1 & 0\\
        \Delta y & -\Delta x & 0 & 0 & 0 & 1
    \end{bmatrix}
    \cdot
    \begin{bmatrix}
        P_x \\
        P_y \\
        P_z \\
        {}^{S}\tau_{0,x} \\
        {}^{S}\tau_{0,y} \\
        {}^{S}\tau_{0,z} 
    \end{bmatrix}
    \label{eq:T_no_cont_sys2}
\end{equation}

\noindent where $y$ is the sensor torque measurement vector, $C = \begin{bmatrix} -S(\Delta \boldsymbol{f}) & I \end{bmatrix}$ is the measurement matrix, $S$ is the skew-symmetric matrix of the position vector $\Delta \boldsymbol{f}$, and $x$ is the vector containing the instrument center of gravity ${}^{S}_{G}\hat{\boldsymbol{p}}$ and the sensor torque bias ${}^{S}\boldsymbol{\tau}_{0}$. So, (\ref{eq:T_no_cont_sys2}) can be solved using the RLS method, similarly to the force compensation model. After estimating ${}^{S}_{G}\hat{\boldsymbol{p}}$ and ${}^{S}\boldsymbol{\tau}_{0}$, the actual external torque ${}^{S}\boldsymbol{\tau}_{contact}$ can be estimated using (\ref{eq:T_cont}). The complete proposed approach is summarized in Algorithm~\ref{alg:combined}. 

\begin{algorithm}[t!]
\caption{Proposed Real-time Compensation Method}
\label{alg:combined}
\begin{algorithmic}[1]
\Require Sensor measurements (${}^{S}\boldsymbol{f}$, ${}^{S}\boldsymbol{\tau}$) and rotation ${}^{E}_{B}\boldsymbol{R}$
\Ensure Estimated parameters (${}^{B}\boldsymbol{f}$, ${}^{S}\boldsymbol{f}_0$, ${}^{S}_{G}\boldsymbol{p}$, ${}^{S}\boldsymbol{\tau}_0$) 

\State Initialize estimates randomly: $\hat{x}^{(f)}_0$, $\hat{x}^{(\tau)}_0$
\State Initialize covariances: $P^{(f)}_0$, $P^{(\tau)}_0 \gets 10^{6}\,I_{6\times 6}$
\State Set measurement noise: $R^{(f)}$, $R^{(\tau)} \gets 2.5\times10^{-3}$
\State Set convergence threshold: $\epsilon \gets 10^{-3}$
\State Initialize flags: $\text{converged}^{(f)}$, $\text{converged}^{(\tau)} \gets \texttt{False}$
\State Define contact thresholds: $f_{th}$, $\tau_{th}$

\While{True}
    \State Acquire sensor measurements ${}^{S}\boldsymbol{f}, {}^{S}\boldsymbol{\tau}$ 
    \State Calculate ${}^{E}_{B}\boldsymbol{R}$ from forward kinematics, ${}^{E}_{B}\boldsymbol{R} \gets ({}^{B}_{E}\boldsymbol{R})^T$
    \State $C^{(f)}_k \gets \begin{bmatrix} {}^{E}_{B}\boldsymbol{R} & I \end{bmatrix}$, \quad $y^{(f)}_k \gets {}^{S}\boldsymbol{f}$
    \State Compute residual norm $r^{(f)} \gets \left\lVert C^{(f)}_k\hat{x}^{(f)}_{k-1} - y^{(f)}_k \right\rVert_2$
    \If{$r^{(f)} < f_{th}$ \textbf{or not} $\text{converged}^{(f)}$}
        \State Compute gain $K^{(f)}_k$ by (\ref{eq:K_k}) 
        \State Update estimate $\hat{x}_{k}^{(f)}$ by (\ref{eq:x_hat}) 
        \State Update error covariance $P^{(f)}_k$ by (\ref{eq:P_k})
        \State $\Delta \hat{x}^{(f)} \gets \hat{x}^{(f)}_{k} - \hat{x}^{(f)}_{k-1}$
        \State $\text{converged}^{(f)} \gets \left (\|\Delta \hat{x}^{(f)}\|_2 < \epsilon \right)$
    \EndIf

    \State Extract ${}^{B}\boldsymbol{f}, {}^{S}\boldsymbol{f}_0$ from $\hat{x}^{(f)}_k$
    \State Compute ${}^{S}\boldsymbol{f}_{contact}$ via (\ref{eq:F_cont})
    \vspace{1em}
    \If{$\text{converged}^{(f)}$}
        \State $\Delta \boldsymbol{f} \gets {}^{S}\boldsymbol{f} - {}^{S}\boldsymbol{f}_0$
        \State $C^{(\tau)}_k \gets \begin{bmatrix} -S(\Delta \boldsymbol{f}) & I \end{bmatrix}$, \quad $y^{(\tau)}_k \gets {}^{S}\boldsymbol{\tau}$
        \State Compute residual norm $r^{(\tau)} \gets \left\lVert C^{(\tau)}_k\hat{x}^{(\tau)}_{k-1} - y^{(\tau)}_k\right\rVert_2$
       \If{$r^{(\tau)} < \tau_{th}$ \textbf{or not} $\text{converged}^{(\tau)}$}
            \State Compute gain $K^{(\tau)}_k$ by (\ref{eq:K_k}) 
            \State Update estimate $\hat{x}^{(\tau)}_{k}$ by (\ref{eq:x_hat}) 
            \State Update error covariance $P^{(\tau)}_k$ by (\ref{eq:P_k})
            \State $\Delta \hat{x}^{(\tau)} \gets \hat{x}^{(\tau)}_{k} - \hat{x}^{(\tau)}_{k-1}$
            \State $\text{converged}^{(\tau)} \gets \left (\|\Delta \hat{x}^{(\tau)}\|_2 < \epsilon \right)$
        \EndIf
    \EndIf

    \State Extract ${}^{S}_{G}\boldsymbol{p}, {}^{S}\boldsymbol{\tau}_0$ from $\hat{x}^{(\tau)}_k$ 
    \State Compute ${}^{S}\boldsymbol{\tau}_{contact}$ via (\ref{eq:T_cont})     
\EndWhile

\Return ${}^{B}\boldsymbol{f}$, ${}^{S}\boldsymbol{f}_0$, ${}^{S}_{G}\boldsymbol{p}$, ${}^{S}\boldsymbol{\tau}_0$
\end{algorithmic}
\end{algorithm}

The proposed approach operates in a cascaded manner. It first compensates for non-contact forces and then for non-contact torques, all in real time. This eliminates the need for dataset collection or repeated calibration. Moreover, estimating the actual external contact force and torque does not require prior knowledge of the robot installation offset ${}^{B}_{G}\boldsymbol{R}$ or the end-effector gravity $mg$. However, these can also be estimated using ${}^{B}\boldsymbol{f}$ as detailed in \cite{huang2024offsets}.

\section{EXPERIMENTAL INVESTIGATION}
\label{sec:Results}

In this section, an experimental investigation is conducted to evaluate the performance of the proposed compensation method. A set of experiments was carried out to verify its effectiveness and robustness in compensating for non-contact forces and torques. 

\subsection{Parameter Identification Under No Contact Force}

\begin{figure}[t!]
\centering
\includegraphics[scale=0.65]{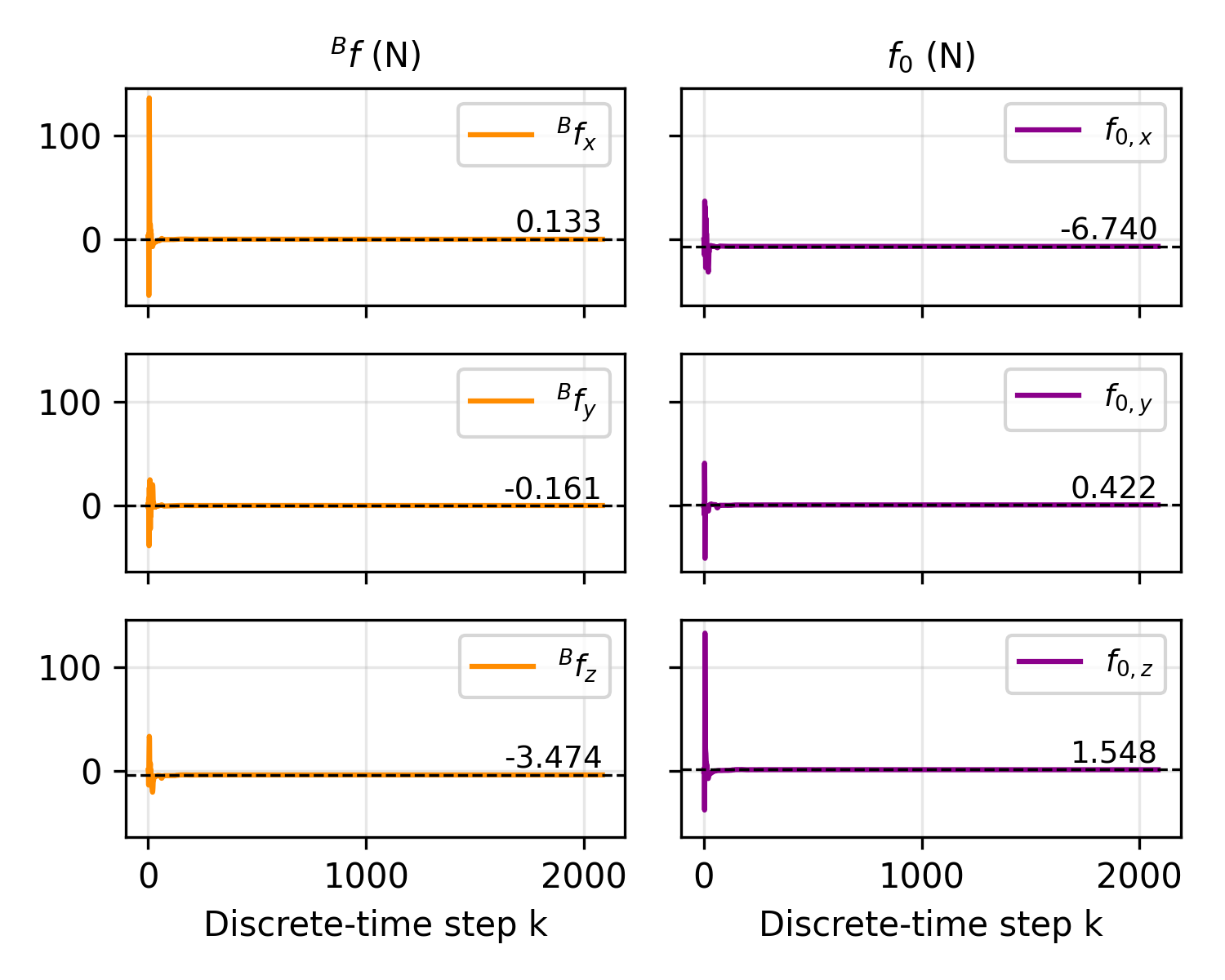}
\caption{Estimation of the force model parameters. Convergence was achieved after 242 measurements, with ${}^{B}\boldsymbol{f}$ on the left and ${}^{S}\boldsymbol{f}_{0}$ on the right.}
\label{fig:Params_force}
\end{figure}

\begin{figure}[t!]
\centering
\includegraphics[scale=0.65]{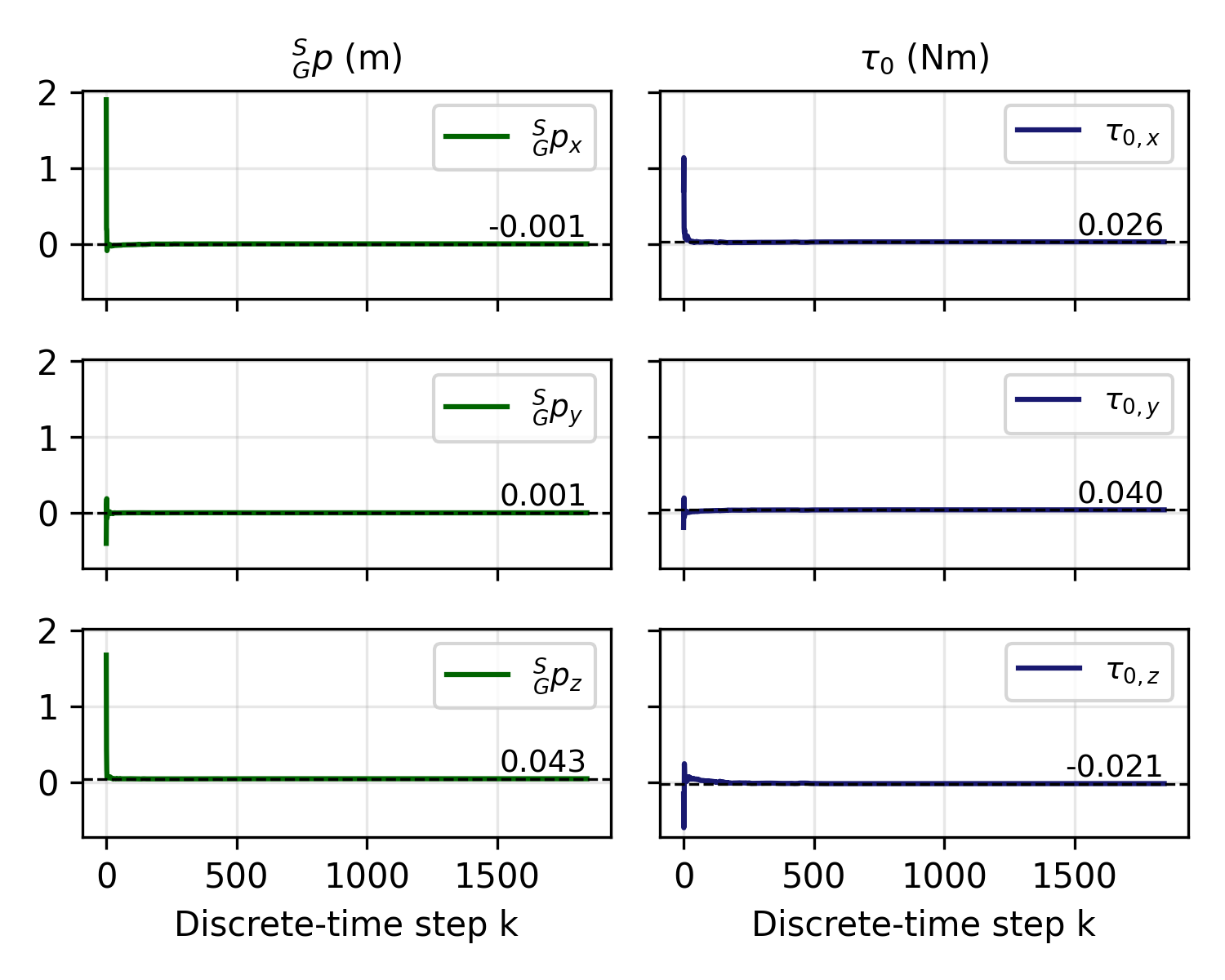}
\caption{Estimation of the torque model parameters. Convergence was achieved after 238 measurements, with ${}^{S}_{G}\boldsymbol{p}$ on the left and ${}^{S}\boldsymbol{\tau}_{0}$ on the right.}
\label{fig:Params_torque}
\end{figure}

In this phase, the unknown model parameters are identified. The robot was moved to random orientations under no external contact, and the corresponding force and torque measurements were recorded while implementing the RLS method in real time, as described in Algorithm~\ref{alg:combined}. The estimation of the unknown force and torque parameters (${}^{B}\boldsymbol{f}$, ${}^{S}\boldsymbol{f}_{0}$, ${}^{S}_{G}\boldsymbol{p}$, ${}^{S}\boldsymbol{\tau}_{0}$) is illustrated in Fig.~\ref{fig:Params_force} and Fig.~\ref{fig:Params_torque}, respectively.

Since the algorithm operates in a cascaded structure, convergence of the force model parameters (${}^{B}\boldsymbol{f}$, ${}^{S}\boldsymbol{f}_{0}$) is required before estimating the torque model parameters (${}^{S}_{G}\boldsymbol{p}$, ${}^{S}\boldsymbol{\tau}_{0}$). In practice, convergence of the force parameters was achieved within only 242 samples. Given the sensor sampling rate of 1~kHz, this ensures a negligible delay in the estimation of the torque parameters.

It is worth noting that the proposed method accurately identified the parameters, particularly the instrument's center of gravity, ${}^{S}_{G}\hat{\boldsymbol{p}}$. The estimated position along the sensor-frame $z$-axis was 0.043 m (Fig.~\ref{fig:Params_torque}), which closely matches the manually measured ground-truth distance of 4.35 cm.


\subsection{Compensation Effectiveness Under No Contact Force}

After identifying the model parameters, the effectiveness of the compensation method was evaluated. In this experiment, the instrument was moved to a series of poses while the rotation matrix of the forward kinematics and the corresponding raw F/T sensor readings were simultaneously recorded. Under the condition of no external forces acting on the instrument, 150 random measurements were collected. 

Since no external contact force was applied, the estimated external wrench ${}^{S}\boldsymbol{f}_{contact}$ and ${}^{S}\boldsymbol{\tau}_{contact}$, defined in (\ref{eq:F_cont}) and (\ref{eq:T_cont}) should ideally be zero. Consequently, the compensation error is characterized by this estimated wrench, which captures the residual difference between the raw sensor wrench and the predicted wrench after compensation.  

\begin{figure}[t!]
\centering
\includegraphics[scale=0.65]{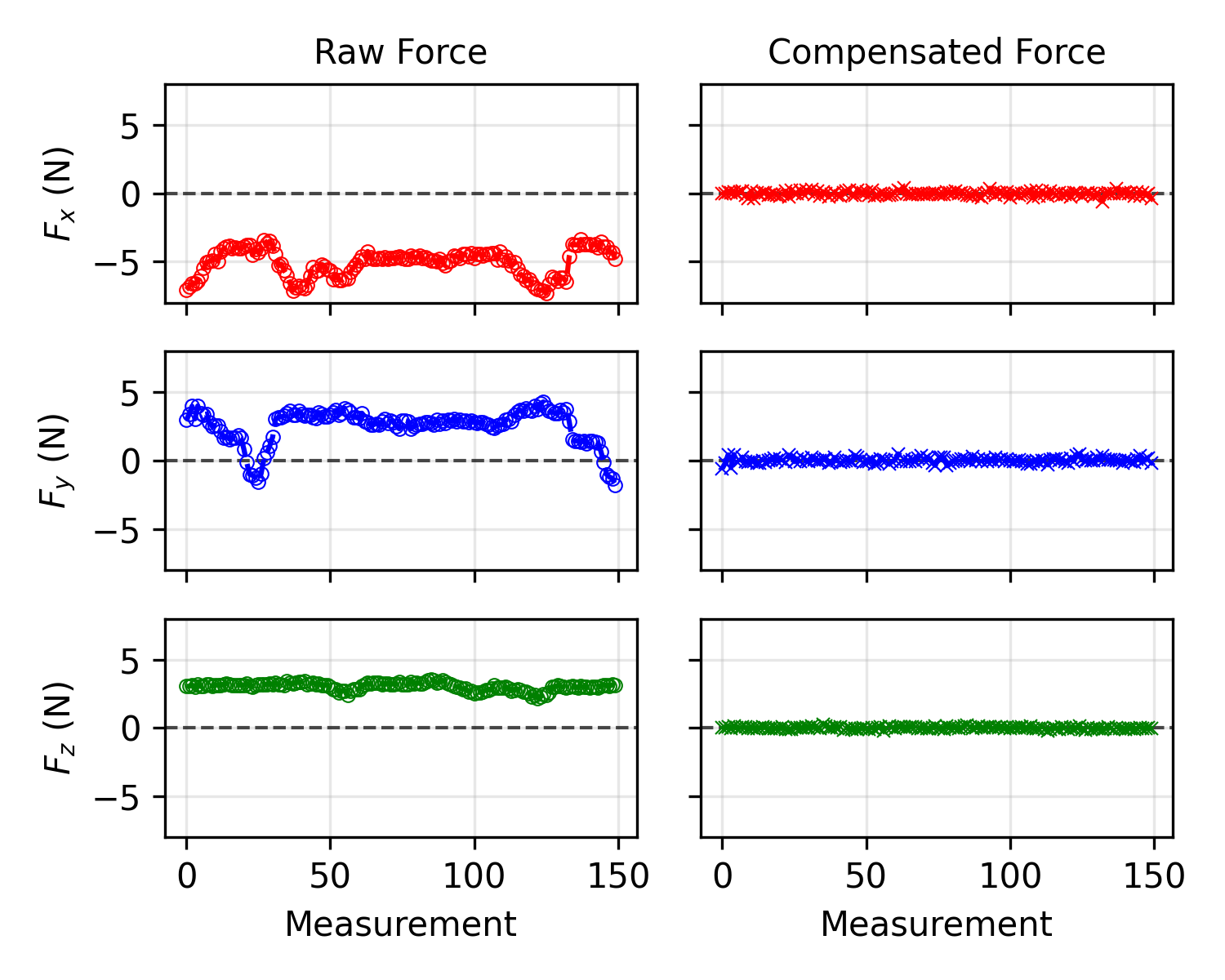}
\caption{Force measurement before and after compensation. The left is the raw sensor data, and the right is the compensated force signals.}
\label{fig:Comp_force}
\end{figure}

\begin{figure}[t!]
\centering
\includegraphics[scale=0.65]{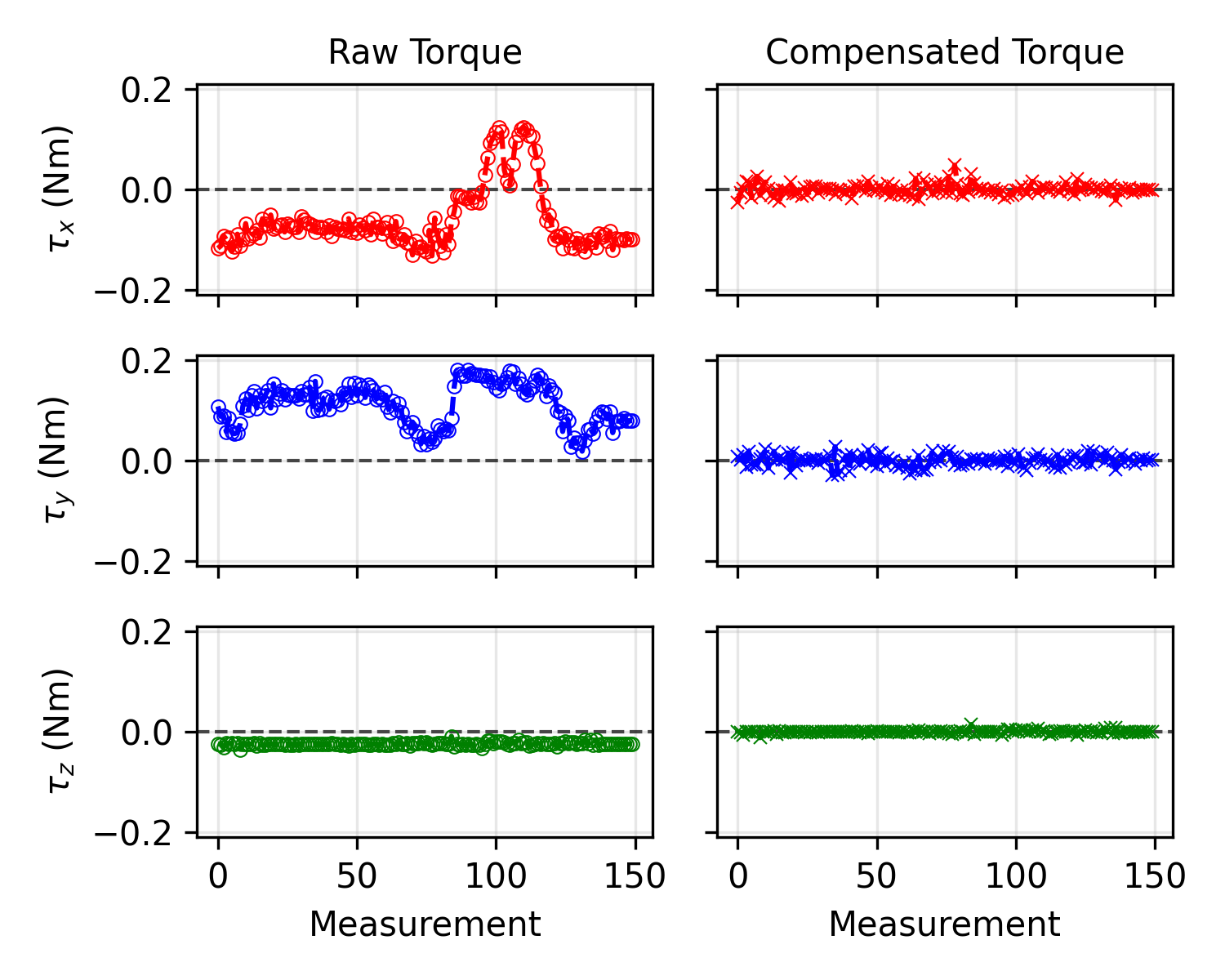}
\caption{Torque measurement before and after compensation. The left is the raw sensor data, and the right is the compensated torque signals.}
\label{fig:Comp_torque}
\end{figure}

As shown in Fig.~\ref{fig:Comp_force} and Fig.~\ref{fig:Comp_torque}, the proposed compensation method significantly reduces the compensation error boundaries across all force and torque components. To further evaluate its effectiveness, the compensation error was analyzed before and after applying the proposed method using three metrics: mean absolute error (MAE) for accuracy reflection, maximum absolute error (Max AE) for worst-case error observation, and standard deviation (Std) for consistency indication. 

Table~\ref{table:Force} and Table~\ref{table:Torque} summarize the error statistics of the force and torque measurements before and after applying the proposed compensation method. For the force components, the MAE decreased from several newtons in the raw data to values close to zero, corresponding to more than 95\% error compensation. Similarly, the torque components exhibited substantial improvements, with MAE compensation above 91\%. These findings demonstrate that the proposed method effectively reduces the error magnitudes and compensates for non-contact forces and torques, thereby enabling more accurate estimation of pure external contact forces.

\renewcommand{\arraystretch}{1.3}
\begin{table}[t!]
\centering
\caption{Force Error Statistics before and after Compensation.}
\begin{tabular}{c c c c c c}
\hline
Component & Condition & MAE & Max AE & Std & MAE Comp. \\
\hline
\multirow{2}{*}{$\boldsymbol{f}_x$} 
 & Before & 5.055 & 7.317 & 0.998 & \multirow{2}{*}{97.8\%} \\
 & After & 0.113 & 0.578 & 0.151 &  \\
\hline
\multirow{2}{*}{$\boldsymbol{f}_y$} 
 & Before & 2.702 & 4.302 & 1.271 & \multirow{2}{*}{95.3\%} \\
 & After & 0.127 & 0.573 & 0.172 &  \\
\hline
\multirow{2}{*}{$\boldsymbol{f}_z$} 
 & Before & 3.049 & 3.530 & 0.262 & \multirow{2}{*}{98.1\%} \\
 & After & 0.059 & 0.254 & 0.076 &  \\
\hline
\end{tabular}
\label{table:Force}
\end{table}

\renewcommand{\arraystretch}{1.3}
\begin{table}[t!]
\centering
\caption{Torque Error Statistics before and after Compensation.}
\begin{tabular}{c c c c c c}
\hline
Component & Condition & MAE & Max AE & Std & MAE Comp. \\
\hline
\multirow{2}{*}{$\boldsymbol{\tau}_x$} 
 & Before & 0.082 & 0.132 & 0.064 & \multirow{2}{*}{91.4\%} \\
 & After & 0.007 & 0.048 & 0.010 &  \\
\hline
\multirow{2}{*}{$\boldsymbol{\tau}_y$} 
 & Before & 0.112 & 0.181 & 0.041 & \multirow{2}{*}{92.8\%} \\
 & After & 0.008 & 0.029 & 0.011 &  \\
\hline
\multirow{2}{*}{$\boldsymbol{\tau}_z$} 
 & Before & 0.024 & 0.035 & 0.003 & \multirow{2}{*}{94.0\%} \\
 & After & 0.001 & 0.015 & 0.003 &  \\
\hline
\end{tabular}
\label{table:Torque}
\end{table}

The performance of the proposed compensation method was benchmarked against the approach presented in \cite{yu2021bias} by comparing the resulting compensation error bounds, as summarized in Table~\ref{table:Comparison}. In \cite{yu2021bias}, the same experiment was conducted across 150 random measurements, yielding force and torque compensation errors bounded within approximately ±1.26~N and ±0.08~N·m, respectively. In contrast, the proposed method achieves substantially tighter error bounds, with worst-case force compensation errors below ±0.58~N and torque errors below ±0.05~N·m. These results demonstrate a clear improvement in compensation accuracy, highlighting the effectiveness of the proposed approach relative to the existing method.

\renewcommand{\arraystretch}{1.3}
\begin{table}[t!]
\centering
\caption{Comparison of Force and Torque Compensation Error Boundaries Between \cite{yu2021bias} and the Proposed Method.}
\begin{tabular}{c c c}
\hline
Component & Method \cite{yu2021bias} & Proposed Method \\
\hline
$\boldsymbol{f}_x$ (N) & ±1.259 & ±0.578 \\
$\boldsymbol{f}_y$ (N) & ±1.122 & ±0.573 \\
$\boldsymbol{f}_z$ (N) & ±1.260 & ±0.254 \\
$\boldsymbol{\tau}_x$ (N·m) & ±0.078 & ±0.048 \\
$\boldsymbol{\tau}_y$ (N·m) & ±0.084 & ±0.029 \\
$\boldsymbol{\tau}_z$ (N·m) & ±0.079 & ±0.015 \\
\hline
\end{tabular}
\label{table:Comparison}
\end{table}

\subsection{Compensation Effectiveness Under External Load}

\begin{figure}[t!]
\centering
\includegraphics[scale=0.55]{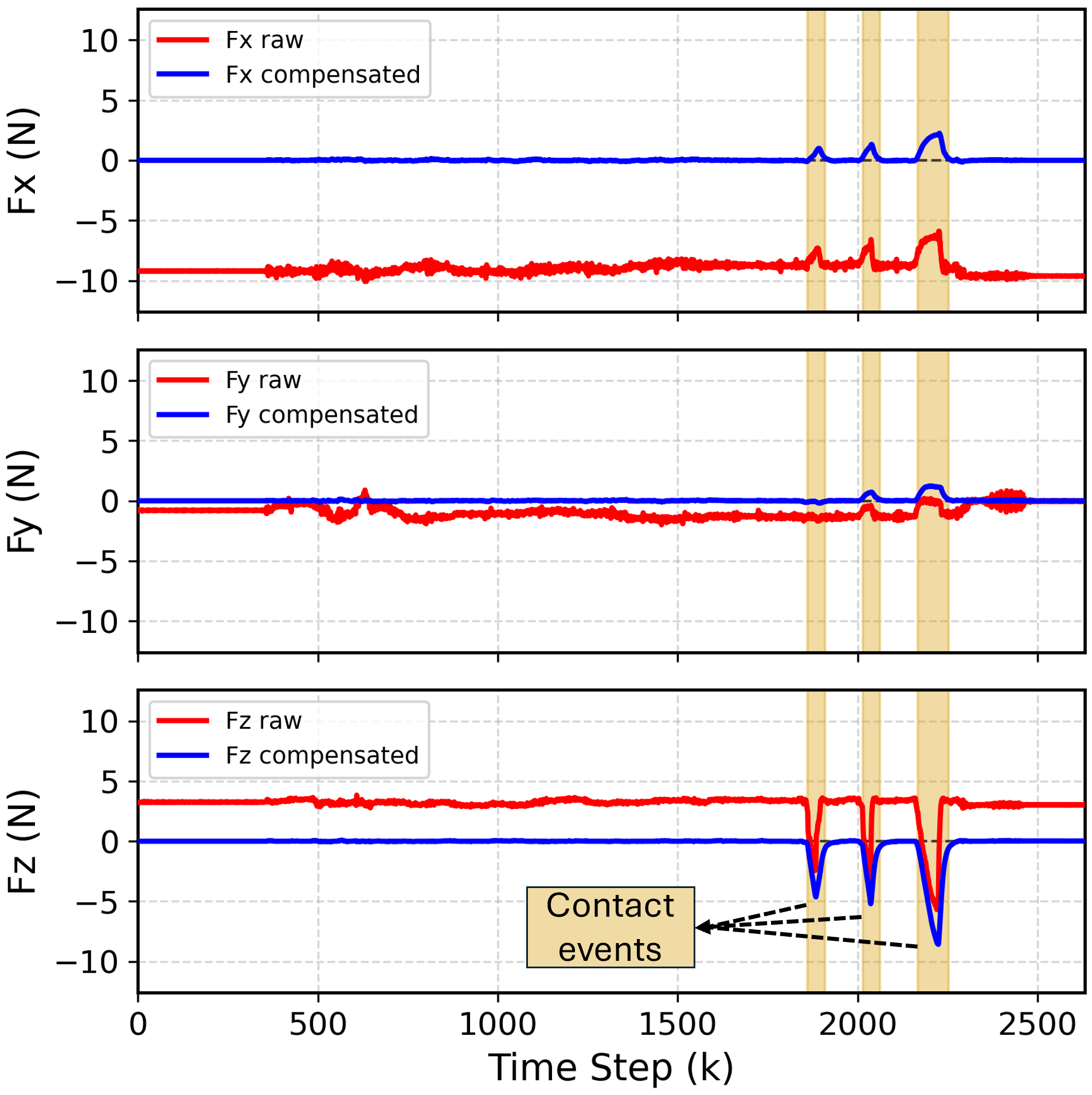}
\caption{Compensation performance under external contact forces. Red curves represent the raw force measurements from the sensor, while blue curves show the compensated forces obtained using the proposed method.}
\label{fig:Contact_force}
\end{figure}

The effectiveness of the proposed compensation method was further evaluated under external contact. In this experiment, the instrument tip repeatedly indented a silicone phantom to generate controlled contact forces. Figure~\ref{fig:Contact_force} illustrates the performance of the proposed compensation method, where the red curves represent the raw force measurements obtained directly from the sensor and the blue curves show the corresponding compensated force components. As observed, the compensation successfully eliminates sensor bias and gravitational effects, ensuring that the force estimates return to zero when no external interaction is present.

Notably, in the regions beyond 1860 samples, where the instrument tip interacts with the phantom, the external contact forces are correctly detected. The compensated signals (blue curves) deviate from zero only during these contact events, accurately reflecting the applied interaction forces while maintaining a smooth transition between contact and non-contact phases.

\begin{figure}[t!]
\centering
\includegraphics[scale=0.69]{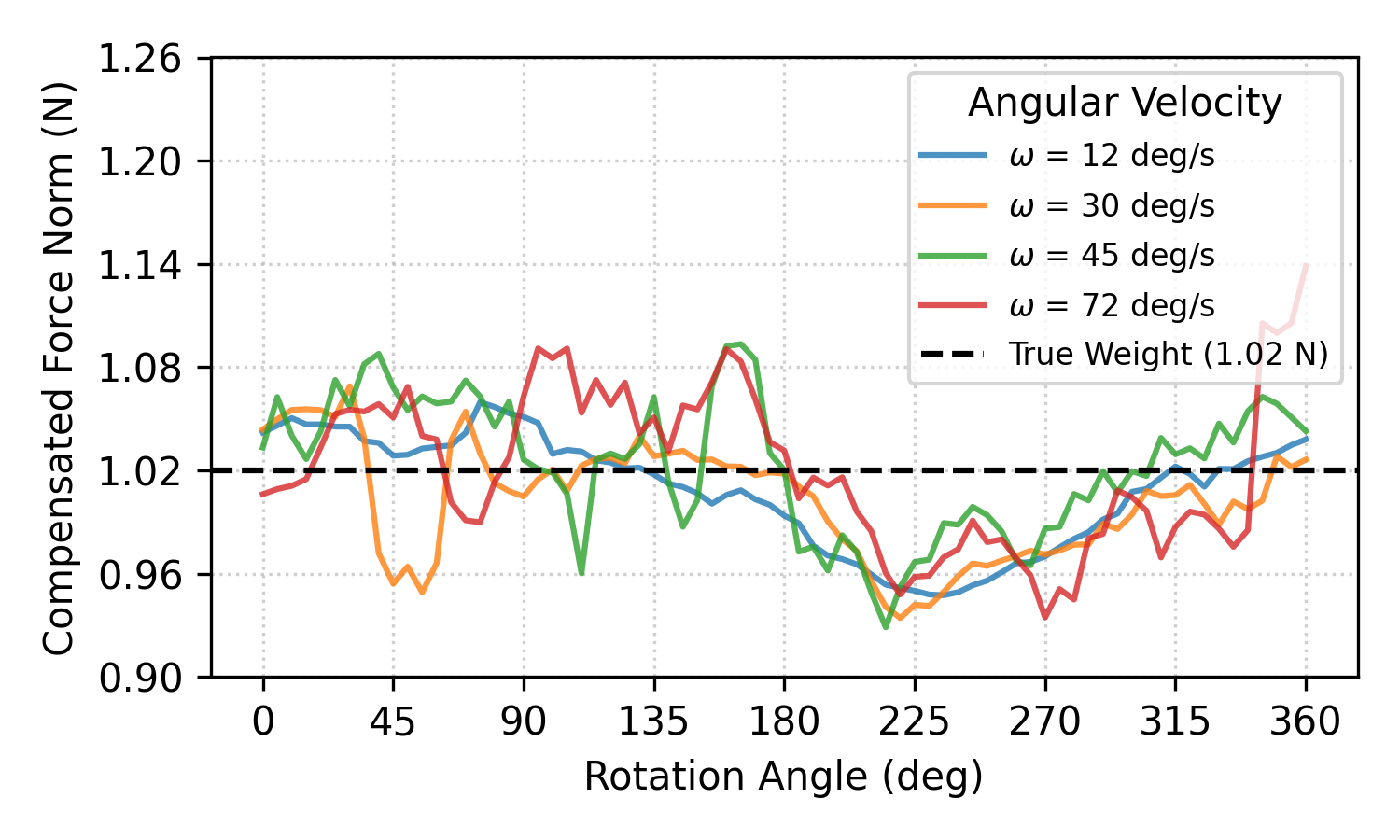}
\caption{Compensated force magnitude during a 360-degree axial rotation of the instrument with a 104g ($\approx$1.02 N) reference mass across four different angular velocities.}
\label{fig:Rotation_360_104g}
\end{figure}

To validate the accuracy of these compensated forces against a known ground truth, another experiment was conducted using a 104g reference mass (corresponding to a gravitational force of $\approx$1.02 N). The instrument is held horizontally and the mass was attached to the tip to simulate a constant lateral contact force while the instrument was rotated 360 degrees around its axis at various angular velocities ($\omega$ = 12, 30, 45, and 72 deg/s), as shown in Fig.~\ref{fig:Rotation_360_104g}. 

Results demonstrate that the compensated force magnitude consistently matched the true load, with a grand mean measurement of 1.014 N, regardless of the sensor's rotation about its z-axis. This behavior confirms the algorithm's ability to maintain accurate real-time compensation even when the direction of gravity changes relative to the sensor during contact.

\begin{figure}[t!]
\centering
\includegraphics[scale=0.69]{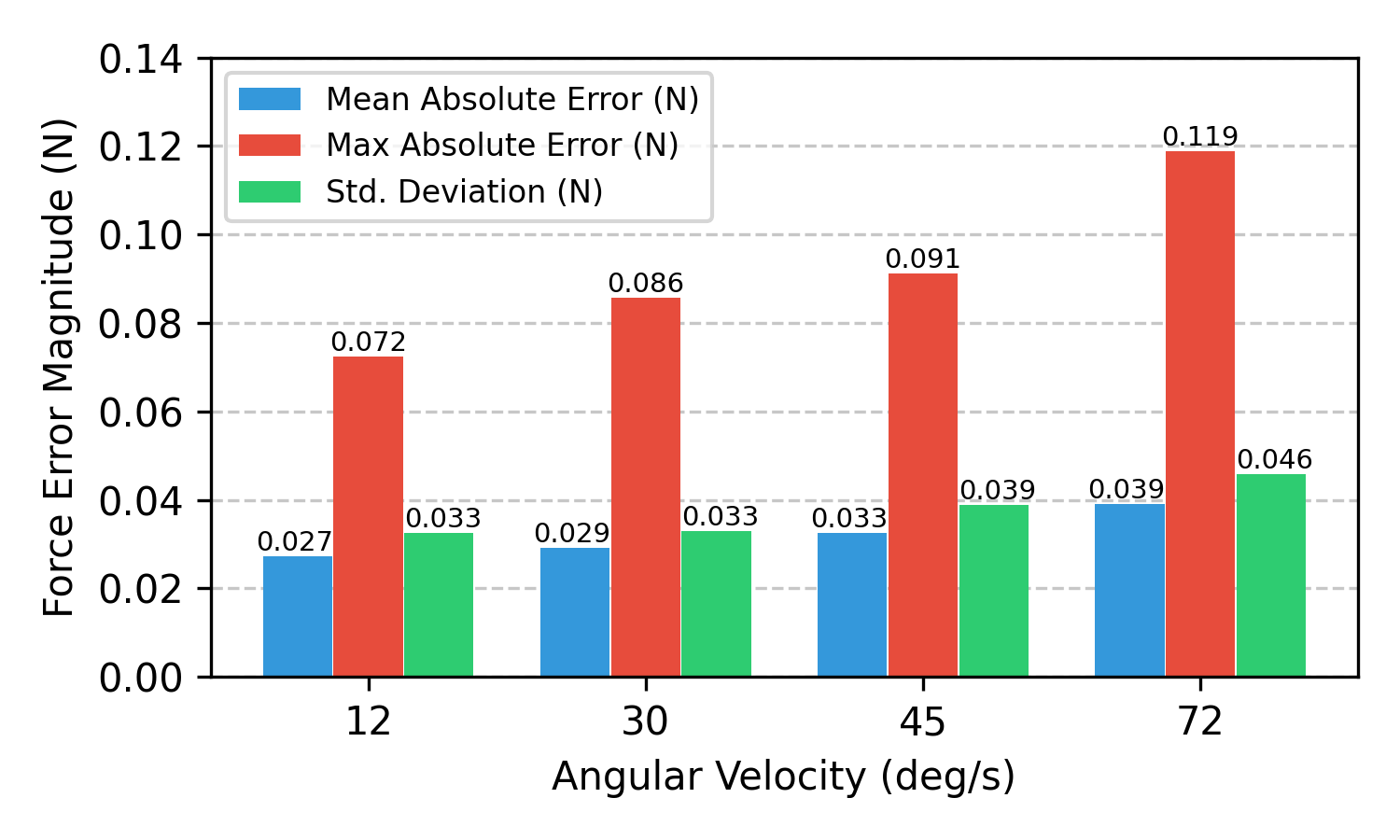}
\caption{Performance evaluation metrics for the tested angular velocities. The MAE and Std remain consistently low across all speeds, while the Max AE shows a slight increase at higher velocities due to inertial effects.}
\label{fig:Rotation_Metrics}
\end{figure}

Figure~\ref{fig:Rotation_Metrics} provides a detailed performance evaluation of the compensation method across the tested angular velocities. The accuracy of the compensated force during interaction was validated using the MAE which remained minimal at low speeds and reached only 0.04 N at the maximum velocity of 72 deg/s, representing a relative error of approximately 3.9\% of the total load. This confirms that the algorithm accurately extracts external contact forces by successfully isolating the instrument's gravity and sensor bias. 

The worst-case error was also analyzed. As shown in Fig.~\ref{fig:Rotation_Metrics}, Max AE generally increases with velocity. This peak error is likely attributable to small inertial transients during faster movements, which the static gravity compensation model does not explicitly account for. However, even the worst-case error (0.119 N at 72 deg/s) remains quite small for haptic feedback applications, which could be investigated further in future work.

The Std remained extremely low at $<$0.046 N across all speeds, highlighting the stability of the compensation method. During the longest test at $\omega$ = 12 deg/s, where contact was maintained for 30 seconds rotation, the compensated force magnitude remained stable with a Std of 0.033 N, indicating no significant drift or instability over time. 

These results demonstrate that the proposed compensation method not only maintains accuracy in the absence of contact but also enables a reliable extraction of the true interaction forces with robustness to both orientation changes and various angular velocities. Such capability is essential for haptic feedback applications, as it ensures that only externally applied forces are conveyed to the user, thereby improving transparency and realism in teleoperation scenarios.

In summary, the proposed compensation method demonstrates superior performance compared to existing approaches. It achieves higher compensation accuracy than state-of-the-art methods when the sensor installation offset is known, while importantly not requiring prior knowledge of the robot installation angle. Unlike conventional techniques, the method operates in real time, eliminating the need for time-consuming data collection and frequent recalibration, while remaining robust to sensor drift and different robot installation configurations. Moreover, the proposed approach can be broadly applicable to force-sensing robotic systems operating in low speeds such as haptic interfaces, surface finishing and human–robot interaction. 

\section{CONCLUSION}
\label{sec:Conc}

This paper presented a robust method to compensate for non-contact forces and torques in wrist-mounted F/T sensors, with the goal of enhancing haptic feedback in robotic surgery training. By employing an RLS approach, the method adaptively compensates for sensor bias, gravitational effects, and installation offsets in real time, without requiring prior calibration or extensive datasets. Experimental results demonstrated substantial improvements in force and torque accuracy, reducing errors by more than 95\% and 91\%, respectively, and yielding tighter error bounds compared to existing methods. Furthermore, interaction tests with a silicone phantom and rotational sweep experiments with a known reference mass demonstrate that the proposed method maintains high fidelity during active external contact. The findings confirm the effectiveness of the proposed approach in isolating true external contact forces, a critical requirement for realistic and transparent haptic feedback. Future work will include investigating faster convergence strategies (e.g., better initialization) for latency-sensitive applications such as compliance control, adaptive thresholding for contact detection to generalize across tools and speeds, and evaluation of adaptation behavior under mid-session payload or tool changes. This will be further extended to surgical tasks and integrated into teleoperated training scenarios, thereby advancing the development of accessible and reliable haptic-enabled robotic surgery training.

\addtolength{\textheight}{-12cm}  


\bibliographystyle{IEEEtran}
\bibliography{ref}
\end{document}